\documentclass[10pt,twocolumn,letterpaper]{article}

\usepackage{cvpr}
\usepackage{times,enumerate}
\usepackage{epsfig}
\usepackage{graphicx}
\usepackage{amsmath}
\usepackage{amssymb}
\usepackage{multirow}
\usepackage{color,balance}
\usepackage{algpseudocode}
\usepackage{algorithm,algorithmicx}

\usepackage[pagebackref=true,breaklinks=true,letterpaper=true,colorlinks,bookmarks=false]{hyperref}

\cvprfinalcopy % *** Uncomment this line for the final submission

\ifcvprfinal\fi
\begin{document}

%%%%%%%%% TITLE
%\title{Deep Extreme Learning Machines based Representation Learning for \\Efficient Image Set Classification}
\title{Representation Learning with Deep Extreme Learning Machines\\ for Efficient Image Set Classification}
\author{Muhammad Uzair$^1$, Faisal Shafait$^1$, Bernard Ghanem$^2$ and  Ajmal Mian$^1$\\
$^{1}$Computer Science \& Software Engineering, The University of Western Australia\\
$^2$ King Abdullah University of Science and Technology, Saudi Arabia\\
{\tt\small muhammad.uzair@research.uwa.edu.au, \{faisal.shafait, ajmal.mian\}@uwa.edu.au } \\ {\tt\small bernard.ghanem@kaust.edu.sa}\\
}

\maketitle
%\thispagestyle{empty}
%\F{Replace MLELM with Deep ELM} \U{I have replaced ML-ELM with DELM}
%%%%%%%%% ABSTRACT

\begin{abstract}

Efficient and accurate joint representation of a collection of images, that belong to the same class, is a major research challenge for practical image set classification. Existing methods either make prior assumptions about the data structure, or perform heavy computations to learn structure from the data itself. In this paper, we propose an efficient image set representation that does not make any prior assumptions about the structure of the underlying data. We learn the non-linear structure of image sets with Deep Extreme Learning Machines (DELM) that are very efficient and generalize well even on a limited number of training samples. Extensive experiments on a broad range of public datasets for image set classification (Honda/UCSD, CMU Mobo, YouTube Celebrities, Celebrity-1000, ETH-80) show that the proposed algorithm consistently outperforms state-of-the-art image set classification methods both in terms of speed and accuracy.
%Compared to the latest deep learning based method, our algorithm reduces training time from about two hours to one second.

\end{abstract}
\vspace{-3mm}
%%%%%%%%%%%%%%%%%%%%%%%%%%%%%%%%%%%%%%%%
%% Introduction
%%%%%%%%%%%%%%%%%%%%%%%%%%%%%%%%%%%%%%%%
\section{Introduction}
Image set based classification has attracted significant interest from the computer vision and pattern recognition community due to its wide range of applications in multi-view object classification \cite{DCC,MMD,MDA,CDL,GGDA,SANSCVPR13} and face recognition \cite{AHISD,SANP,HSSC,AjmalTIP,LMKML,NLRM,ECCV14}. Image set classification naturally arises in many applications when a given collection of images are known to belong to one class but with unknown identity.
In contrast to the traditional paradigm of single image based classification, algorithms for image set classification exploit this information to obtain a more accurate estimate of the class identity. Multiple images of a set usually contain a range of intra-class appearance variations such as pose, illumination and scale changes, which can be explicitly or implicitly modelled for improved classification accuracy~\cite{SANP,DCC,NLRM,LMKML}. Image set based classification may also be considered as a generalization of video based object classification. However, image set classification does not assume any temporal relationship between the images that constitute the set. Thus, image set classification is also applicable in situations where the set samples have large variations without any temporal relationship~\cite{SANP,CDL}.

\begin{figure*}
\vspace{-3mm}
\begin{center}
   \includegraphics[width = 15cm]{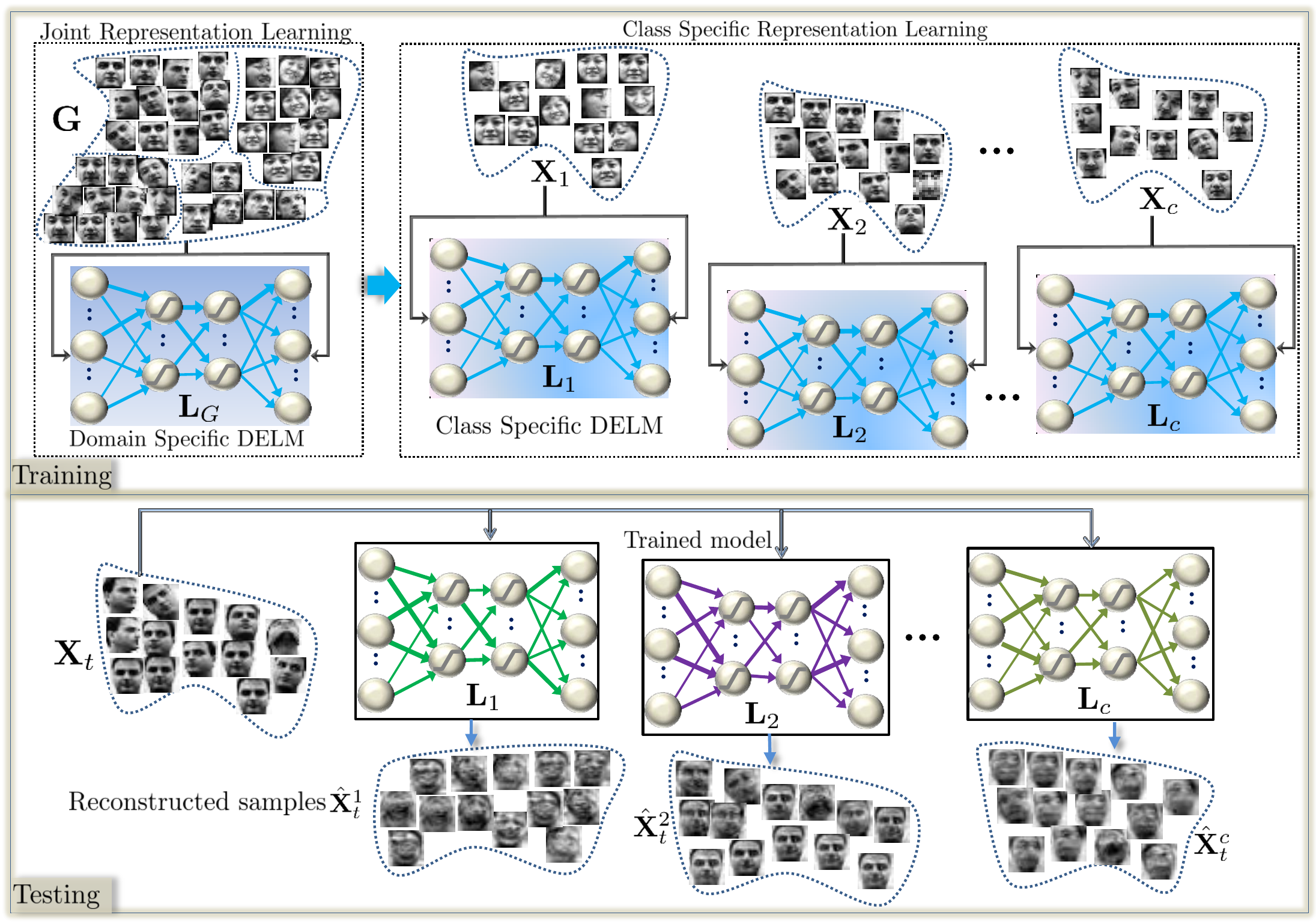}
\end{center}
\vspace{-3mm}
\caption{Illustration of the proposed algorithm. During training, we first learn a domain-specific Deep Extreme Learning Machines (DELM) model $\mathbf{L}_G$. Starting from the domain specific model, we then learn class-specific DELM models $\mathbf{L}_j$ for the gallery sets of each class separately. Given a probe image set $\mathbf{X}_t$, we first reconstruct each of its samples using the learned DELM models and then estimate its label based on the smallest reconstruction error. Finally, majority voting is used to estimate the label of the probe set as a whole.  }
\label{fig:mainfig}
\vspace{-3mm}
\end{figure*}
An image set classification algorithm must essentially address two core challenges; how to represent an image set to effectively capture intra-image as well as inter-image variations and how to define a distance/similarity measure between two image sets. Defining a suitable distance between two image sets is often tied to the representation used to model the image sets in the first place. Hence, most of the research in this area has concentrated on developing image set representations based on certain assumptions about the set structure. Some techniques assume that the set data follows a Gaussian distribution~\cite{MDA,CDL,LMKML,UZICB} which is unlikely to be true for all types of images. Other methods assume that an image set can be represented by linear subspaces~\cite{DCC,GGDA}, whereas the data may lie on complex manifolds~\cite{NLRM}. To model more complex data structures, several techniques have been proposed to model image sets as a convex or affine hulls of the data samples~\cite{AHISD,SANP,ECCV14}. These techniques are conceptually similar to nearest neighbor classification and must impose certain constraints to avoid finding the neighbors in some low dimensional space where image sets might intersect. However, the ability to model more complex image set structures comes at the cost of added algorithm complexity~\cite{MMD, MDA, SANP,LMKML,ECCV14,NLRM}. Therefore, these algorithms cannot be efficiently scaled to handle large image set classification tasks \cite{Celeb1000}.

In this work, we have focused on developing an efficient and accurate representation of image sets that can model arbitrarily complex image set structures on one hand, and scale to large problem sizes on the other. We employ Extreme Learning Machines (ELM) for this purpose primarily due to their computational efficiency~\cite{ELMNC,ELMTNN,ELMSMCB2012,ELMSMCB2014,ELMP1}. An ELM trains a single hidden layer feed-forward neural network (SLFN) by randomly assigning weights to the input layer and analytically computing the weights for the output layer. Deep ELMs have the potential of effectively learning the underlying structure of the image set without any prior assumption on the distribution or structure of image set data. Our algorithm learns a Deep ELM (DELM) model for each class in the gallery (training classes) through unsupervised feature learning with an ELM based auto-encoder (ELM-AE) (Fig.~\ref{fig:mainfig}). A label is assigned to a probe (test) set based on minimum reconstruction error.

The key contributions of this paper are three-fold: (1) An effective image set representation scheme based on Deep Extreme Learning Machines that does not make any assumption about the structure of the set but implicitly learns it from training data. (2) Unlike existing deep learning based methods, our algorithm does not require a large amount of training data. (3) The proposed network is extremely fast both in training and test -- training is 6,000 times faster than the state-of-the-art best-performing method, whereas the testing is 9 times faster. The proposed algorithm is extensively evaluated for image set based face recognition and object categorization on five benchmark datasets including Honda/UCSD~\cite{HONDA}, CMU Mobo~\cite{MOBO},  YouTube Celebrities~\cite{Youtube}, Celebrity-1000~\cite{Celeb1000} and ETH-80 \cite{ETH}. Results demonstrate that our algorithm consistently outperforms existing methods in terms of accuracy, while achieving substantial speedups at the same time.

%%%%%%%%%%%%%%%%%%%%%%%%%%%%%%%%%%%%%%%%
%% Related Work
%%%%%%%%%%%%%%%%%%%%%%%%%%%%%%%%%%%%%%%%
\vspace{-3mm}
\section{Related Work}
\vspace{-2mm}
Existing image set classification techniques can be categorized into sample based and structure based set-to-set matching methods. Sample based techniques compute the distance between the nearest neighbors of two image sets under certain constraints. For example, Cevikalp and Triggs~\cite{AHISD} model each image set as a convex geometric region in feature space. Set dissimilarity between the regions represented by the affine (AHISD) or convex hulls (CHISD) is measured by the distance of closest point approach. For the affine hull model, the distance is minimized using least squares while for the convex hull model, an SVM is trained to separate the two sets.  Hu \etal~\cite{SANP} approximate each of the two nearest points between two image sets by a sparse combination of the corresponding set samples. The sparse approximated nearest points (SANP) lie close to some facets of the affine hulls and hence, implicitly incorporate structural information while matching two sets. To find more accurate nearest points between two image sets, Mian \etal \cite{AjmalTIP} introduced self-regularized non-negative coding to define between set distance. They constrained the orthogonal basis vectors to be similar to the approximated nearest points and added the non-negativity constraint on the set samples while approximating nearest points. Mahmood \etal\cite{HSSC} performed spectral clustering on the combined gallery and test samples. The class-cluster distributions of the set samples were then used for classification. Lu \etal \cite{ECCV14} jointly learn a structured dictionary and projection matrix to map set samples into a low-dimensional subspace. The low dimensional samples are then represented using sparse codes and classification is performed based on the traditional minimum reconstruction error and majority voting scheme. In general, sample based methods are highly prone to outliers and are computationally expensive for large galleries.

Structure based techniques model an image set with one or more linear subspaces. Structural similarity is then measured using a subspace-to-subspace distance. Kim \etal~\cite{DCC} perform discriminant analysis on the canonical correlations calculated between set structures. Wang \etal \cite{MMD} model an image set with multiple local clusters and represent each cluster with a linear subspace. Subspace distance between the nearest local clusters of two sets is then used for classification. Chen \etal \cite{SANSCVPR13} proposed sparse approximated nearest subspaces (SANS) to extract local linear models from the gallery image sets via sparse representation. By forcing the clusters of the query image set to resemble clusters in the gallery image sets, only corresponding clusters are matched using the subspace based distance. Wang and Chen ~\cite{MDA} proposed Manifold Discriminant Analysis (MDA) which models each image set using multiple local linear clusters. These clusters are transformed by a linear discriminant operator to separate different classes. Here, the set-to-set similarity is measured using pair-wise local cluster distances in the learned embedding space. Harandi \etal \cite{GGDA} modeled the image set structure with linear subspaces as points lying on Grassmannian manifolds. They define kernels to map points from the Grassmannian manifold to Euclidean space where classification is performed by graph embedding discriminant analysis. Wang \etal \cite{CDL} model the structure of each image set directly using a covariance matrix. They map the covariance matrix of each image set from the Riemannian manifold to the Euclidean space by a kernel function based on the Log Euclidean distance. Image sets are then classified according to a learned regression function using Kernel Partial Least Squares. Hayat \etal \cite{NLRM} learn the structure of each gallery image set using a deep learning model. The label of the test set is then estimated based on the minimum reconstruction error and majority voting scheme. Generally, structure based algorithms require a relatively large number of images in each set (dense sampling) in order to accurately model the underlying structure.

We propose a structure based image set classification algorithm that neither makes prior assumptions about the set structure nor incur a heavy computational burden to learn the structure from the data. The proposed representation is based on deep Extreme Learning Machines and automatically learns the {\em non-linear} structure of image sets. The proposed algorithm is extremely efficient to train and generalizes very well even with a limited number of training samples.

%\B{We need to discuss the main differences with this other work. This section reads like a list of papers with not much relation to the proposed method.} \U{Please check the newly added paragraph.}

%%%%%%%%%%%%%%%%%%%%%%%%%%%%%%%%%%%%%%%%
%% Extreme Learning Machines (Overview)
%%%%%%%%%%%%%%%%%%%%%%%%%%%%%%%%%%%%%%%%
\section{Proposed Methodology}
We first give a brief overview of Extreme Learning Machines (ELMs) and how they differ from other learning paradigms. Then, we discuss how to extend the traditional ELM idea to multiple layers, thus, allowing a deeper representation. Finally, we show how image set classification can be formulated using the Deep ELM (DELM) models and how it can benefit from ELM's attractive properties, namely very efficient learning (easily scalable to large datasets) and generalizability (no prior assumptions on the set data).

% overview of ELM
\subsection{Extreme Learning Machines}
Consider a supervised learning problem with $N$ training samples, $\{\mathbf{X},\mathbf{T}\} = \{\mathbf{x}_j,\mathbf{t}_j\}_{j=1}^N$ where $\mathbf{x}_j \in \mathbb{R}^d$ and $\mathbf{t}_j \in \mathbb{R}^q$ are the $j^{\text{th}}$ input and target samples respectively. $d$ and $q$ are the input and target feature dimensions respectively. For the task of classification, $\mathbf{t}_j$ is the class label vector while for regression $\mathbf{t}_j$ represents the desired output feature. In either case, we seek a regressor function from the inputs to the targets. One popular form of this function is the standard single hidden layer feed-forward network (SLFN), where $n_h$ hidden nodes fully connect the $d$ inputs to the $q$ outputs. This is done through an activation function $g(u)$. The predicted output vector $\mathbf{o}_j$ generated by feeding forward $\mathbf{x}_j$ through an SLFN is mathematically modelled as
%\begin{equation}
\begin{align}
    %\mathbf{o}_j=\sum_{i=1}^{n_h} \boldsymbol{\beta}_i g_i(
    \mathbf{o}_j=\sum_{i=1}^{n_h} \boldsymbol{\beta}_i g(\mathbf{w}_i^{\top}  \mathbf{x}_j + b_i)
\end{align}
%\end{equation}
\noindent where $\mathbf{w}_i \in \mathbb{R}^{d}$ is the weight vector connecting the $i$-th hidden node and the input nodes, $\boldsymbol{\beta}_i \in \mathbb{R}^q$ is the weight vector connecting the $i$-th hidden node and the output nodes, and $b_i$ is the bias of the $i$-th hidden node. The activation function $g(u)$ can be any non-linear piecewise continuous function, such as the sigmoid function $g(u) = \frac{1}{1 + e^{-u}}$.

An ELM learns the parameters of an SLFN (i.e. $\{\mathbf{w}_i,b_i,\boldsymbol{\beta}_i\}_{i=1}^{n_h}$) in two sequential stages: random feature mapping and linear parameter solving. In the first ELM stage, the hidden layer parameters ($\{\mathbf{w}_i,b_i\}_{i=1}^{n_h}$) are randomly initialized to project the input data into a random ELM feature space using the the mapping function $g(.)$. It is this random projection stage that differentiates ELM from most existing learning paradigms, which perform deterministic feature mapping. For example, an SVM uses kernel functions, while deep neural networks \cite{Bengio2009} use Restricted Boltzmann machines (RBM) for feature mapping/learning. By randomizing the feature mapping stage, the ELM can discover non-linear structures in the data without the need for priors, which are inherently the case for deterministic feature mapping schemes. Also, these parameters are set randomly and are not subsequently updated, thus decoupling them from the output parameters $\{ \boldsymbol{\beta}_i\}_{i=1}^{n_h}$, which can be learned in a very efficient manner as we will see next. This decoupling strategy significantly speeds up the parameter learning process in ELM, thus, making it much more computationally attractive than deep neural network architectures that learn \emph{all} network parameters \emph{iteratively}. %s\B{Relate this to work done on random projections in sparse coding. See Yi Ma's PAMI paper on the topic of face recognition.}

In the second ELM stage, the parameters connecting the hidden layer and the output layer (\ie $\{ \boldsymbol{\beta}_i\}_{i=1}^{n_h}$) are learned efficiently using regularized least squares. Here, we denote
$\psi(\mathbf{x}_j) = [ g(\mathbf{w}_1^{\top}  \mathbf{x}_j + b_1) \dots g(\mathbf{w}_{n_h}^{\top}  \mathbf{x}_j + b_{n_h})] \in \mathbb{R}^{1 \times n_h}$
as the response vector of the hidden layer to the input $\mathbf{x}_j$ and $\mathbf{B} \in \mathbb{R}^{n_h \times q}$ as the output parameters connecting the hidden and output layers. An ELM aims to solve for $\mathbf{B}$ by minimizing the sum of the squared losses of the prediction errors:
\vspace{-3mm}
\begin{align}
& \underset{\mathbf{B} \in \mathbb{R}^{n_h \times q}}{\min} \frac{1}{2} \|\mathbf{B}\|_F^2 + \frac{C}{2} \sum_{j=1}^N \|\mathbf{e}_j\|_2^2 \label{eq:ELM1}\\
& s.t.~~~~~ \psi(\mathbf{x}_j)\mathbf{B} = \mathbf{t}_j^{\top} - \mathbf{e}_j^{\top}, ~~~j = 1,...,N \nonumber
\end{align}
In (\ref{eq:ELM1}), the first term is a regularizer against over-fitting, $\mathbf{e}_j \in \mathbb{R}^q$ is the error vector with respect to the $j$-th training pattern (i.e. $\mathbf{e}_j = \mathbf{t}_j - \mathbf{o}_j$), and $C$ is a tradeoff coefficient. By concatenating $\mathbf{H}=[ \psi(\mathbf{x}_1)^{\top} \cdots \psi(\mathbf{x}_N)^{\top}]^{\top } \in \mathbb{R}^{N \times n_h}$ and $\mathbf{T}=[\mathbf{t}_1\cdots\mathbf{t}_N]^{\top} \in \mathbb{R} ^ {N \times q}$, we obtain an equivalent unconstrained optimization problem, which is widely known as ridge regression or regularized least squares.
\begin{equation}
\underset{\mathbf{B} \in \mathbb{R}^{n_h \times q}}{\min} \frac{1}{2} \|\mathbf{B}\|_F^2 + \frac{C}{2} \|\mathbf{T} - \mathbf{H}\mathbf{B} \|_2^2,
\label{eq:ELM2}
\end{equation}
%where $\mathbf{H}=[ g(\mathbf{x}_1)^{\top}, . . ., g(\mathbf{x}_N)^{\top}]^{\top} \in \mathcal{R}^{N\times n_h}$
Since the above problem is convex, its global solution needs to satisfy the following linear system.
\begin{equation}
\mathbf{B} + C \mathbf{H}^{\top}(\mathbf{T} - \mathbf{H}\mathbf{B}) = \mathbf{0}.
\label{eq:ELMRLS}
\vspace{-3mm}
\end{equation}
The solution to this system depends on the nature and size of matrix  $\mathbf{H}$. If $\mathbf{H}$ has more rows than columns and is of full column rank (which usually is the case when $N>n_h$), the system is overdetermined and a closed form solution exists for (\ref{eq:ELM2}) in (\ref{eq:ELMRSOL}), where $\mathbf{I}_{n_h}\mathbb{R}^{n_h \times n_h}$ is an identity matrix. Note that in practice, rather than explicitly inverting the $n_h \times n_h$ matrix, we obtain $\mathbf{B}^{\ast}$ by solving the linear system in a more efficient and numerically stable manner.
\begin{equation}
\mathbf{B}^{\ast} = \biggl ( \mathbf{H}^{\top} \mathbf{H} + \frac{\mathbf{I}_{n_h}}{C} \biggr)^{-1} \mathbf{H}^{\top} \mathbf{T}
\label{eq:ELMRSOL}
\vspace{-3mm}
\end{equation}
If $N<n_h$, $\mathbf{H}$ will have more columns than rows, which often leads to an under-determined least
squares problem. In this case, $\mathbf{B}$ may have infinite number of solutions. In this case, we restrict $\mathbf{B}$ to be a linear combination of the rows of $\mathbf{H}: \mathbf{B} = \mathbf{H}^{\top} \boldsymbol{\alpha} ~~(\boldsymbol{\alpha} \in \mathbb{R}^{N\times q})$. Note that when $\mathbf{H}$ has more columns than rows and is of
full row rank, then $\mathbf{H}\mathbf{H}^{\top}$ is invertible. Multiplying both sides of (\ref{eq:ELMRLS}) by $(\mathbf{H}\mathbf{H}^{\top})^{-1}\mathbf{H}$, we obtain a closed form solution for $\mathbf{B}^*$
\begin{equation}
\mathbf{B}^{\ast} = \mathbf{H}^{\top} \boldsymbol{\alpha}^{\ast} = \mathbf{H}^{\top} \biggl ( \mathbf{H} \mathbf{H}^{\top} + \frac{\mathbf{I_{N}}}{C} \biggr)^{-1} \mathbf{T}
\label{eq:ELMFINALSol}
\vspace{-2mm}
\end{equation}
To summarize, ELMs have two notably attractive features. Firstly, the parameters of the hidden mapping function can be randomly generated according to any continuous probability distribution \eg the uniform distribution on $[-1, 1]$. Secondly, as such, the only parameters to be learned in training are the output weights between the hidden and output nodes. This can be done by solving a single linear system or even in closed form. These two features make ELMs more flexible than SVMs and much more computationally attractive than traditional feed-forward neural networks that use back-propagation \cite{ELMTNN}.

% motivate multi-layer ELM (or ML-ELM) for auto-encoding
%%%%%%%%%%%%%%%%%%%%%%%%%%%%%%%%%%%%%%%%
%% Learning with ELMs
%%%%%%%%%%%%%%%%%%%%%%%%%%%%%%%%%%%%%%%%
\subsection{Learning Representations with ELMs} \label{sec:ELMAE}

\begin{figure}
\begin{center}
   \includegraphics[width = 8.5cm]{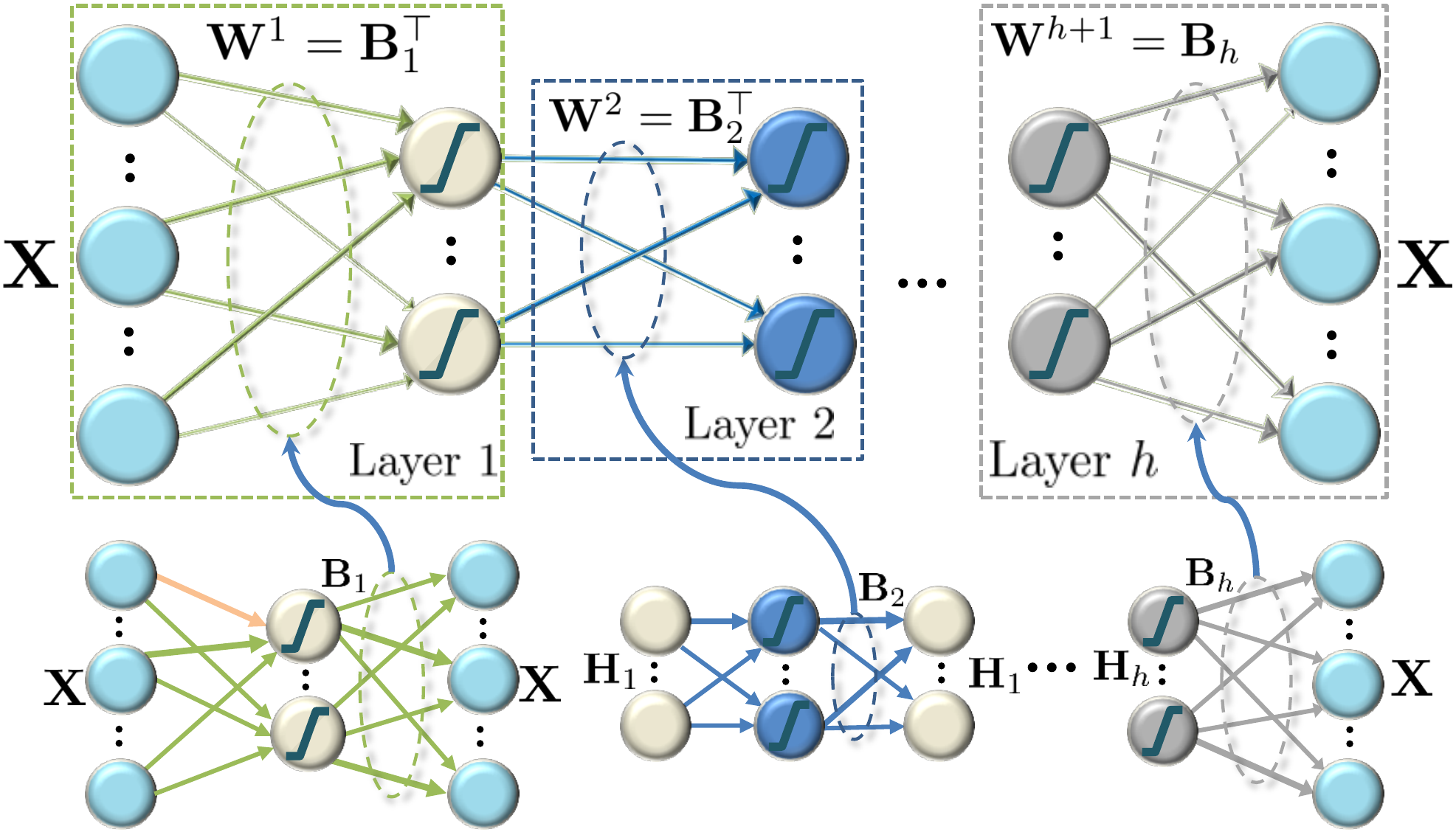}
\vspace{-3mm}
\end{center}
\vspace{-4mm}
\caption{Layer wise training of a Deep ELM model with $h$ hidden layers and input $\mathbf{X}$}
\label{fig:ELMAE}
\vspace{-4mm}
\end{figure}

Learning rich representations efficiently is crucial for achieving high generalization performance, especially at large scales. This form of learning can usually be done using stacked auto-encoders (SAE) and stacked auto-decoders (SDA), where a parametric regressor function is learned to map the input to itself. Although deep neural networks can be learned for this purpose and have been shown to yield good performance in various computer vision tasks \cite{Bengio2009,BengioPAMI13}, they are generally very slow in training. To address the problem, we learn representations in an unsupervised way using an ELM based auto-encoder \cite{ELMAE}, which in essence is a multi-layer feed-forward network whose parameters are learned by cascading multiple layers of ELM. This ELM-based learning procedure is highly efficient and has good generalization capabilities.

The deep ELM auto-encoder is designed by setting the targets of the multi-layer network to the input \ie $\mathbf{T} = \mathbf{X}$. Figure \ref{fig:ELMAE} shows the process of learning a DELM model from the samples of the training set $\mathbf{X}$. Here, we consider a fully connected multi-layer network with $h$ hidden layers. Let $\mathbf{L} = \{\mathbf{W}^1,...,\mathbf{W}^{h+1}\}$ denote the parameters of the DELM that need to be learned, where $\mathbf{W}^i=[\mathbf{w}_1^i,...,\mathbf{w}_{n_i}^i]^{\top} \in \mathbb{R}^{n_{i+1}\times n_i}$. To simplify training, each layer is decoupled within the network and processed as an ELM, whose targets are the same as its inputs. As shown in Figure \ref{fig:ELMAE}, $\mathbf{W}^1$ is learned by considering a corresponding ELM with $\mathbf{T} = \mathbf{X}$.

The weight vectors connecting the input layer to each unit of the first hidden layer are orthonormal to each other, effectively resulting in projection of the input data to a random subspace. Compared to initializing random weights independent of each other, orthogonalization of these random weights tends to better preserve pairwise distances in the random ELM feature space~\cite{JL} and improves ELM auto-encoding generalization performance. Next, $\mathbf{B}_1$ is calculated using \eqref{eq:ELMRSOL} or \eqref{eq:ELMFINALSol} depending on the number of nodes in the hidden layer. Note that,  $\mathbf{B}_1$ re-projects the lower dimensional representation of the input data back to its original space while minimizing the reconstruction error. Therefore, this projection matrix is data-driven and hence used as the weights of the first layer ($\mathbf{W}^1 = \mathbf{B}_1^{\top}$). Similarly, $\mathbf{W}^2$ is learned by setting the input and output of Layer 2 to $\mathbf{H}_1$ \ie the output of Layer 1. In this manner, all parameters of the DELM can be computed sequentially. However, when the number of nodes between two consecutive layers is equal, the random projection obtained in the second layer is in the same space as the input of the first layer. Using \ref{eq:ELMRSOL} or \ref{eq:ELMFINALSol} does not ensure orthogonality of the computed weight matrix $\mathbf{B}$. Imposing orthogonality in this case results in a more accurate solution since the data always lies in the same space. Therefore, the output weights $\mathbf{B}$ are calculated as the  solution to the Orthogonal Procrustes problem.
\vspace{-0.8mm}
\begin{align}
\mathbf{B}^{\ast}  = &\underset{\mathbf{B} \in \mathbb{R}^{n_h \times q}}{\min} \|\mathbf{H}\mathbf{B} - \mathbf{T}\|_F^2,\\
&s.t. ~~\mathbf{B}^{\top}\mathbf{B} = \mathbf{I} \nonumber
\end{align}
The close form solution is obtained by  finding the nearest orthogonal matrix to the given matrix $\mathbf{M}=\mathbf{H}^{\top}\mathbf{T}$. To find the orthogonal matrix $\mathbf{B}^{\ast}$, we use the singular value decomposition $\mathbf{M}=\mathbf{U} \boldsymbol{\Sigma} \mathbf{V}^{\top}$ to compute  $\mathbf{B}^{\ast}=\mathbf{U}\mathbf{V}^{\top}$.

In ELM-AE, the orthogonal random weights and biases of the hidden nodes project the input data to a different or equal dimension space. The DELM models can automatically learn the non-linear structure of data in a very efficient manner. In contrast to deep networks, DELM also does not require expensive iterative fine tuning of the weights.

% apply ML-ELM for image set classification
\subsection{Deep ELM models for Image set Classification}
DELM based image set classification is performed in two steps. We first learn a global domain-specific DELM model using all the training image data and then build class-specific DELM models using the global representation as an initialization. In doing so, we encode both domain level and class-specific properties of the data.

Let $\mathbf{G} = \{\mathbf{X}_m\}_{m=1}^{c} \in\mathbb{R}^{d\times N}$ be the gallery containing $c$ image sets of $c$ different classes and $N$ images: $N = \sum_{m=1}^{c}{s_m}$, where $s_m$ is the number of image samples in the $m$-th image set. Let $\mathbf{X}_m=\{\mathbf{x}_m^i\}_{i=1}^{s_m} \in\mathbb{R}^{d\times s_m}$ be the $m$-th image set, where $\mathbf{x}_m^i\in\mathbb{R}^{d}$ is a $d$-dimensional feature vector obtained by vectorizing the pixels of the $i$-th image. Instead of pixel values, the vector $\mathbf{x}_m^i$ may also contain other features, \eg local binary patterns (LBP). While $s_m$ can vary across image sets, the dimensionality of $\mathbf{x}_m^i$ remains fixed. Let $\mathbf{Y }= \{y_m\}_{m=1}^{c}$ be the class labels of the image sets in $\mathbf{G}$. For a test image set $\mathbf{X}_t=\{\mathbf{x}_t^i\}_{i=1}^{s_t} \in\mathbb{R}^{d\times s_t}$, the problem of image set classification involves estimating the label $Y_t$ of $\mathbf{X}_t$ given the gallery $\mathbf{G}$.

\vspace{-3mm} %%%%%%%%%%%%%%%%%%%%%%%%%%%%%
\paragraph{Training:} We learn a global domain-specific DELM model by initializing its weights using the ELM auto-encoding procedure described earlier. This global DELM is a multi-layer neural network with $h$ hidden layers. Its parameters are learned using the images in $\mathbf{G}$ in an unsupervised manner.  The global DELM model is represented as $\mathbf{L}_G = \{\mathbf{W}_G^1,...,\mathbf{W}_G^{h+1}\}$, where $\mathbf{W}_G^i$ is the weight matrix of the $i^{\text{th}}$ layer learned using the auto-encoding method in Section \ref{sec:ELMAE}. The global DELM model serves as a starting point, from which we learn class-specific DELM models.

%\begin{equation}
%\mathbf{L}_G = \{\mathbf{W}_G^1,...,\mathbf{W}_G^{h+1}\}
%\end{equation}

Since $\mathbf{L}_G$ encodes domain-specific representation (as it has been trained to reconstruct any sample from that domain), we use it to learn a separate DELM model for each of the $c$ training classes.
In other words, instead of randomly initializing the hidden layers weights, as in the conventional ELM, we use the weights in $\mathbf{L}_G$ to initialize the class-specific models. Thus, we have $c$ DELM models for $c$ classes $\{\mathbf{L}_j\}_{j=1}^c$, where each class-specific model is represented as $\mathbf{L}_j = \{\mathbf{W}_j^1,...,\mathbf{W}_j^{h+1}\}$.
%\B{Motivate why this is better than just simply randomly initializing the class-specific models. } \F{Motivation given at the start of the paragraph}

%\begin{equation}
%\mathbf{L}_c = \{\mathbf{W}_c^1,...,\mathbf{W}_c^{h+1}\}
%\end{equation}

The learned ELM models are able to encode complex non-linear structure of the training data due to their deep architecture with multiple non-linear layers. Compared to the previous structure based algorithms such as DCC \cite{DCC}, GGDA \cite{GGDA} and CDL \cite{CDL}, our proposed DELM models learn the structure of the image data in multiple parameters, therefore, it is capable of learning more complex non-linear manifold structures. Moreover, this DELM model is more computationally efficient than previous methods.

\vspace{-3mm}
\paragraph{Testing:} Given a test image set $\mathbf{X}_t=\{\mathbf{x}_t^i\}_{i=1}^{s_t}$, we predict its label by first representing each image in this set using each of the class-specific representations $\{\mathbf{L}_j\}_{j=1}^c$ and assigning each image to the class that incurs the least reconstruction error. Then, majority voting on the predicted image-level classes is performed to predict the class of the image set. The overall procedure is summarized in Algorithm \ref{algo:main_algorithm}.

\begin{algorithm}[!tb]
\caption{Proposed Image Set Classification Algorithm}
\begin{algorithmic}
\Require:
\State Gallery $\mathbf{G}=\{\mathbf{X}_m\}_{m=1}^{c}$ containing $c$ image sets~$\mathbf{X}_s=\{\mathbf{x}_m^i\}_{i=1}^{s_m} \in\mathbb{R}^{d\times s_m}$ belonging to $c$ classes
\State Class labels $\mathbf{Y }= \{y_m\}_{m=1}^{c}$
\State Probe set $\mathbf{X}_t=\{\mathbf{x}_t^i\}_{i=1}^{s_t} \in\mathbb{R}^{d\times s_t}$
\State Number of hidden layers $h$
\Ensure: Label $Y_t$ of $\mathbf{X}_t$
\State \textbf{Training:}
\State $\mathbf{L}_G = \{\mathbf{W}_G^1,...,\mathbf{W}_G^{h+1}\}$ \{Learn a domain-specific global DELM model with $h$ hidden layers from $\mathbf{G}$\}
\For {$j = 1:c$}
\State $\mathbf{L}_j = \{\mathbf{W}_j^1, ... \mathbf{W}_j^{h+1} \}$ \{ Learn DELM models for each class\}
\EndFor
\State \textbf{Testing:}
\For {$i = 1:s_t$}
\For {$j = 1:c$}
%\State $\hat{\mathbf{x}}^i_t=g(\{\mathbf{x};\mathbf{W}_c^{h+1}\}g(\{\mathbf{x};\mathbf{W}_c^h\},...,g(\{\mathbf{x};\mathbf{W}_c^1\})))$
\State $\hat{\mathbf{x}}^i_j= f(\mathbf{x}^i_t;\mathbf{L}_j)$\{Reconstruct from model $\mathbf{L}_j$\ (\ref{eq:reconF})\}
\State $e^i(j) = \|\mathbf{x}^i_t-\hat{\mathbf{x}}^i_j\|_2^2$
\EndFor
\State ${l}_t^i \triangleq \underset{j}{\arg\min}~~e^i(j)$
\EndFor
%\State $Y_t \triangleq \text{mode}(\mathbf{l}_t)$ \{ Label estimation for $\mathbf{X}_t$\}
\State $Y_t \triangleq \text{mode}(\{l_t^i\}_{i=1}^{s_t})$
\end{algorithmic}
\label{algo:main_algorithm}
\end{algorithm}

We reconstruct each test image $\mathbf{x}_t^i$ in the set using each of the class-specific models $\{\mathbf{L}_j\}_{j=1}^c$. The reconstructed sample $\hat{\mathbf{x}}_j^i$ from a model $\mathbf{L}_j$ is given by
\vspace{-2mm}
\begin{equation}
\hat{\mathbf{x}}^i_j = f(\mathbf{x}_t^i,\mathbf{L}_j) = g(\mathbf{W}_j^{h+1}g(\mathbf{W}_j^h,...,g(\mathbf{W}_j^1 \mathbf{x}_t^i)))
\label{eq:reconF}
\end{equation}
where $f$ is the reconstruction  and $g$ is chosen to be the sigmoid function. The reconstruction error of sample $\mathbf{x}_t^i$ is computed as the squared Euclidean distance between  $\mathbf{x}_t^i$  and $\hat{\mathbf{x}}^i_j$ as $e^i(j) = \|\mathbf{x}^i_t-\hat{\mathbf{x}}^i_j\|_2$. The predicted label $l_t^i$ for sample $\mathbf{x}_t^i$ is chosen to be the class that incurs the minimum reconstruction error
\vspace{-2mm}
\begin{equation}
{l}_t^i = \underset{j}{\arg\min}~~e^i(j)~.
\vspace{-2mm}
\end{equation}
Finally, the test image set $\mathbf{X}_t$ is labelled using majority voting on the set of predicted image-level labels. Formally, we set the image set label $Y_t=\text{mode}(\{l_t^i\}_{i=1}^{s_t})$.

%%%%%%%%%%%%%%%%%%%%%%%%%%%%%%%%%%%%%%%%%%%%%%%%%%%%%%%%%%%%%%%%
%% Experiments
%%%%%%%%%%%%%%%%%%%%%%%%%%%%%%%%%%%%%%%%%%%%%%%%%%%%%%%%%%%%%%%%
\section{Experimental Results}\label{sec:exp}
\begin{figure*}[t]
\begin{center}
   \includegraphics[width=17 cm]{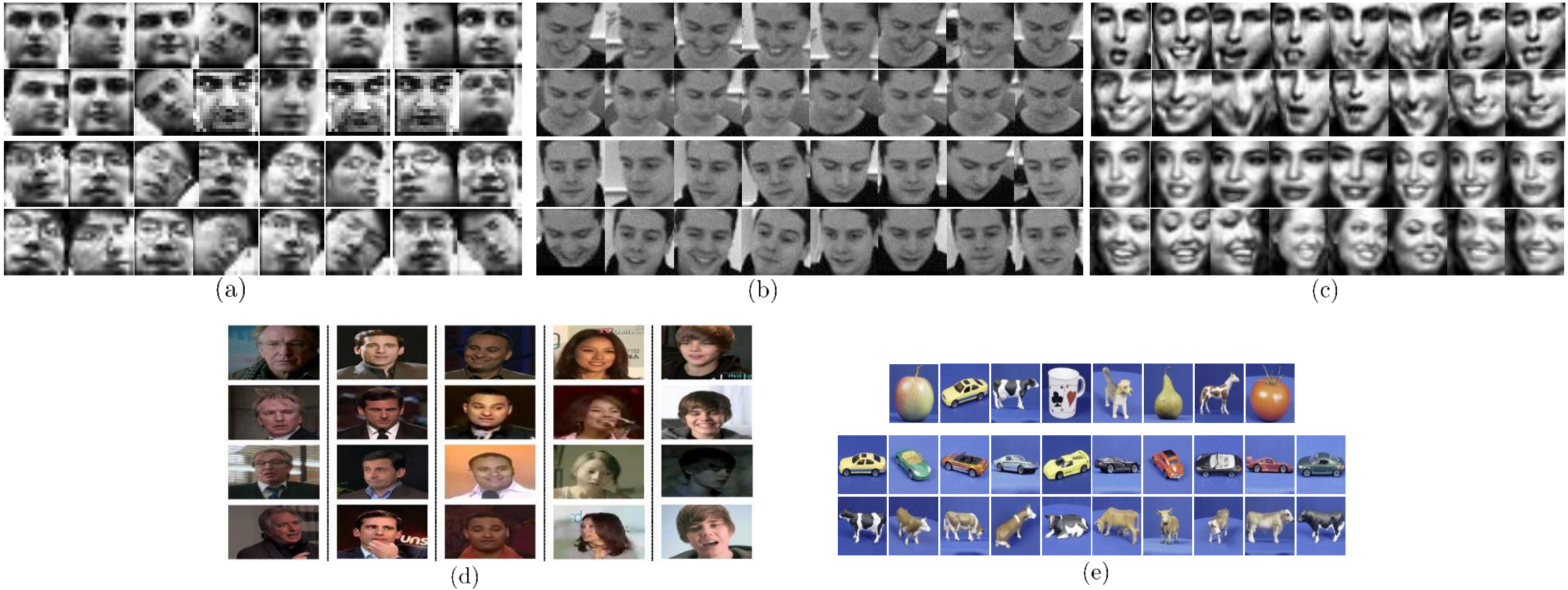}
\end{center}
\vspace{-2mm} \caption{Image sets from (a) Honda, (b) CMU Mobo, (c) Youtube Celebrities, (d) Exemplar video frames from the Celebrity-1000 dataset. (e) 8 Object categories and 10 different objects in one category of the ETH-80 dataset.} \label{fig:datasets}
\vspace{-4mm}
\end{figure*}

We perform extensive experiments on five public datasets (see Fig.~\ref{fig:datasets}) and compare results to 10 state-of-the-art image set classification methods. These datasets have been widely used in
the literature to evaluate image set based classification algorithms. Details of the datasets used, experimental protocol, and results obtained are provided next.

\subsection{Dataset Specifications}

The \textbf{Honda/UCSD dataset}~\cite{HONDA} contains 59 video sequences of 20
different subjects. Similar to prior work~\cite{SANP,CDL}, we use $20 \times 20$ histogram equalized face images extracted from these videos. Each video sequence corresponds to an image set.

The \textbf{CMU MoBo dataset} contains 96 video sequences of 24 different subjects. We use LBP features of the face images as in~\cite{AHISD} for image set classification.

The \textbf{YouTube Celebrities}~\cite{Youtube} is a challenging dataset that contains 1,910 video sequences of 47 celebrities (actors, actresses and politicians), collected from YouTube. Most videos are of low resolution and contain significant compression artifacts. There are upto 400 frames per video. We use the LBP features ($d=928$) of $20 \times 20$ face images.

%The \textbf{Labeled Faces in the Wild (LFW) dataset}~\cite{LFWTech} contains face images captured in unconstrained environments, exhibiting strong variations in pose, lighting, focus, resolution, facial expression, age, gender, race, accessories, make-up, occlusions and background. We use the LFW-a~\cite{LFWA} version of the images which are aligned with a commercial face alignment software. We crop the aligned images to 90$\times$75. We select all 62 subjects, having at least 20 images each and generate disjoint sets by randomly selecting half  images of each subject for training sets and the remaining half as testing set.

The \textbf{Celebrity-1000 database}~\cite{Celeb1000} is a large-scale unconstrained video database downloaded from Youtube and Youku. It contains 159,726 video sequences of 1,000 subjects covering a wide range of poses, illuminations, expressions and image resolutions. We follow the standard closed-set test protocol defined in~\cite{Celeb1000} where four overlapping subsets of the dataset are created with increasing complexity containing 100, 200, 500, 1000 subjects. Each subset is further divided into training and test partitions with disjoint video sequences. Approximately 70\% of the sequences are randomly selected to form the gallery and the rest are used as probes. We use the PCA reduced LBP+Gabor features provided by Liu \etal~\cite{Celeb1000} . The feature dimension $d$ is 1651, 1790, 1815 and 1854 for the subsets 100, 200, 500 and 1000 respectively.

%We report the results of our algorithm alone on the full dataset but limit the comparative evaluation to the smallest sets (200, 100 and 500 subjects) as existing algorithms have a very high computational complexity and memory requirements.

The \textbf{ETH-80 dataset}~\cite{ETH} contains 8 object categories, where each category has 10 different objects of the same class. Each object has 41 images at different views to make an image set. We use 20 $\times$ 20 intensity images for image set based object categorization. ETH-80 is a challenging as it has fewer images per set, significant appearance variations across objects of the same class and larger viewing angle differences within each image set.

\subsection{Experimental Setup}
\label{sec:exp-setup}

We follow the standard experimental protocol \cite{MMD,MDA,AHISD,SANP,CDL,NLRM} for a fair comparison with 10 state of the art algorithms including
Discriminant Canonical Correlation (DCC)~\cite{DCC}, Manifold-Manifold Distance~\cite{MMD}, Manifold Discriminant Analysis (MDA)~\cite{MDA}, Affine and Convex Hull based Image Set Distance (AHISD, CHISD)~\cite{AHISD}, Sparse Approximated Nearest Points (SANP)~\cite{SANP}, Covariance Discriminative Learning (CDL)~\cite{CDL}, Graph Embedding Grassmannian Discriminant Analysis (GGDA)~\cite{GGDA}, Set to Set Distance Metric Learning (SSDML)~\cite{SSDML} and Non-Linear Reconstruction Models (NLRM)~\cite{NLRM}. We use the source codes supplied by the original authors, except for MDA and CDL techniques. For MDA, Hu’s~\cite{SANP} implementation is used, while we use our own implementation of CDL.

Parameters of all the algorithms are selected empirically and the best results are reported. For DCC \cite{DCC}, the subspace dimension and the corresponding maximum canonical correlations is set to 10. For MMD and MDA, we configure the parameters as recommended by the authors \cite{MMD, MDA}. The ratio between Euclidean distance and geodesic distance is selected from the range $\{$1.0-5.0$\}$ for different data sets. The maximum canonical correlation is used in defining MMD. The number of connected nearest neighbors for computing geodesic distance in both MMD and MDA is set to 12. For AHISD, CHISD and SANP, the PCA energy used to represent an image set is selected from the range $\{$80\%, 85\%, 90\%, 95\%, 99\%$\}$ and the best results are reported for each dataset. For CHISD, we set the error penalty parameter $C = 100$. For GGDA, we set $k^{[cc]} = 1$ $k^{[proj]} = 100$ and $v = 3$. The number of eigenvectors used to represent an image set is set to 9 and 6 respectively for Mobo and YouTube Celebrities and 10 for all other datasets. No parameter settings are required for CDL and SSDML. For NLRM~\cite{NLRM}, we used the network depth and model parameters as recommended by the authors. The parameters of our algorithm include the number of hidden layers $h$, the number of neurons in each hidden layer $n_h$ and the parameters $C$. We set the number of hidden layers $h = 2$ for all datasets. The parameter $C$ is in the range $\{ 10^4-10^8\}$ for the first layer and $\{ 10^{16}-10^{20}\}$ for the last layer. The number of neurons in each hidden layer $n_h$ is 20 for Honda, Mobo and Celebrity-1000, 40 for Youtube, 150 for ETH80.

For Honda and MoBo data sets, each subject has one video sequence in the gallery and the rest in probes. For DCC learning, at least two image-sets per class are required in the gallery. Therefore, when the gallery contained only one image-set per class, we randomly partitioned it into two non-overlapping sub-sets. Experiments were repeated 10-folds with different gallery/probe combinations. For Youtube dataset, we follow the experimental protocol of~\cite{SANP} and conduct five-fold cross-validation experiments. The videos are divided to make nine image sets per subject in each fold. In each fold, three image sets per subject are randomly selected for training and the rest are used for testing. For ETH-80 dataset, each class has 5 sets in the gallery for training and the remaining 5 sets are used as probes.

\begin{table*}[t]
\caption{Comparison of the average recognition rates and standard deviations (\%) (Results are obtained by performing 10-fold experiments on Honda, Mobo and ETH datasets and 5-fold on Youtube celeberities dataset.)}
%\vspace{-2mm}
\begin{center}
\begin{tabular}{llccc}
\hline
\hline
                          & Honda          & MoBo            & ETH-80          & Youtube \\
\hline
\hline
    DCC\cite{DCC}         & 94.67$\pm$1.32 & 93.61$\pm$1.76  & 90.91$\pm$5.31  & 66.75$\pm$4.47\\
%\hline
    MMD\cite{MMD}         & 94.87$\pm$1.16 & 93.19$\pm$1.66  & 85.73$\pm$8.33  & 65.12$\pm$4.36 \\
%\hline
    MDA\cite{MDA}         & 97.44$\pm$0.91 & 95.97$\pm$1.90  & 80.50$\pm$6.81  & 68.12$\pm$4.85 \\
%\hline
   GGDA\cite{GGDA}        & 94.61$\pm$2.07 & 85.75$\pm$1.82  & 85.75$\pm$6.41  & 62.81$\pm$4.42\\
%\hline
   CDL\cite{CDL}          & 100.0$\pm$0.00 & 95.83$\pm$2.07  & 88.20$\pm$6.80  & 68.96$\pm$5.29\\
%\hline
   AHISD\cite{AHISD}      & 89.74$\pm$1.85 & 94.58$\pm$2.57  & 74.76$\pm$3.31  & 71.92$\pm$4.55 \\
%\hline
    CHISD\cite{AHISD}     & 92.31$\pm$2.12 & 96.52$\pm$1.18  & 71.00$\pm$3.93  & 73.17$\pm$4.69  \\
%\hline
    SANP\cite{SANP}       & 93.08$\pm$3.43 & 97.08$\pm$1.03  & 72.43$\pm$4.98  & 74.01$\pm$4.68\\
%\hline
   SSDML\cite{SSDML}      & 89.41$\pm$3.64 & 95.14$\pm$2.20  & 81.00$\pm$6.58  & 70.81$\pm$3.42\\
%\hline
   NLRM\cite{NLRM}       & 100.0$\pm$0.0 & 97.92$\pm$1.76    & 95.25$\pm$4.77  & 73.55$\pm$4.74 \\
%\hline
   Proposed DELM       &\textbf{100.0$\pm$0.0} & \textbf{98.00$\pm$0.67}&\textbf{96.00$\pm$3.51}& \textbf{75.31$\pm$4.63}\\
\hline
%\hline
\end{tabular}
\end{center}
\label{tab:results}
%\vspace{-7mm}
\end{table*}

\begin{table*}[t]
\caption{Comparison of the classification accuray on different subsets of Celeb-1000 dataset.}
%\vspace{-2mm}
\begin{center}
\begin{tabular}{llcccr}
\hline
\hline
                          & Subset-100 & Subset-200 & Subset-500 & Subset-1000 & Average \\
\hline
\hline
    DCC\cite{DCC}         & 25.24   & 10.38  &  10.18  & - & -  \\
%\hline
    MMD\cite{MMD}         & 17.52   & 10.23  &  9.79   & - & - \\
%\hline
    MDA\cite{MDA}         & 15.93   & 9.21   &  9.87   & - & - \\
%\hline
   GGDA\cite{GGDA}        & 11.95   & 8.24   &  9.64   & - & - \\
%\hline
   CDL\cite{CDL}          & 11.95   & 11.11  &  10.65  & - & - \\
%\hline
   AHISD\cite{AHISD}      & 19.92   & 23.94  &  18.97  & - & - \\
%\hline
    CHISD\cite{AHISD}     & 20.31   & 22.41  &  18.35  & - & - \\
%\hline
    SANP\cite{SANP}       & 20.71   & 21.64  &  19.12  & - & - \\
%\hline
   SSDML\cite{SSDML}      & 18.32   & 17.62  &  9.96   & - & - \\
%\hline
   NLRM\cite{NLRM}        & 34.66   & 31.81  &  27.68  &  - & - \\
%\hline
	MTJSR\cite{Celeb1000} & \textbf{50.59}   & 40.80  &  35.48  &  \textbf{30.03} & 39.22\\
%\hline
   Proposed DELM          & 49.80     & \textbf{45.21} & \textbf{38.88}& 28.83 & \textbf{40.68} \\
\hline
%\hline
\end{tabular}
\end{center}
\label{tab:resultsCeleb}
%\vspace{-7mm}
\end{table*}

\vspace{-2mm}
\subsection{Results and Analysis}
Table~\ref{tab:results} reports the average and standard deviation recognition rate (\%) for 10-fold experiments on Honda, Mobo and ETH datasets and 5-fold experiments on the Youtube dataset. Our approach performs better than competing algorithms on Youtube celebrities, CMU Mobo and ETH-80 datasets and achieves perfect results on the Honda dataset. %Compared to algorithms that represent  image set structure in a linear fashion (e.g. DCC, MMD and CDL), our DELM method inherently learns a non-linear structure for the data.
Recall that our algorithm involves no supervised discriminative analysis as in DCC, MDA and CDL, yet it performs superior in both accuracy and execution time. On the ETH-80 dataset, structure based algorithms~\cite{DCC, MMD, MDA, NLRM, CDL} perform better than the sample based ones~\cite{AHISD,SANP} because the individual samples can not model significant intra-class pose and object appearance variations. %By capturing non-linear structure in the image-sets efficiently, the proposed DELM achieves the best results on this dataset. On LFW-a dataset, all image set classification algorithms perform well where our algorithm outperforms the others.

Table~\ref{tab:resultsCeleb} summarizes the image set classification results on all the splits of the Celebrity-1000 dataset. On the subset-100 (Celeb-100) our method achieves a 15\% improvement in classification accuracy compared to the existing state-of-the-art. As the feature dimension and dataset size is huge, the training and testing time of all other methods is very large on this dataset (for example on the Celeb-100 the NLRM~\cite{NLRM} method took about 60 hours for training and the MMD and MDA took more than 80 hours using a Core i7 3.4GHz CPU with 8GB RAM). In contrast, our method takes only 5.02 seconds for training and achieves better classification accuracy than all previous methods. Similarly, on the subset-200 the NLRM method took about 5 days for training and the MMD and MDA took more than 8 days. On subset-200 DELM takes only 9.02 seconds for training and achieves better classification accuracy.

The subset-1000 contains 15 million frames in 1000 training image sets and  36 thousands frames in 2580 test image sets. Therefore, previous image set classification methods have a huge computational and memory requirement on this subset. This makes the experimental evaluation and the parameter tuning of these methods very difficult and extremely time consuming. Therefore, on the subset-1000 we only report the results of the proposed algorithm and compare to Multi-Task Joint Sparse Representation (MTJSR)~\cite{Celeb1000}. Note that the accuracies of MTJSR in Table~\ref{tab:resultsCeleb} are provided by the original author ~\cite{Celeb1000}. The proposed algorithm has comparable or better accuracy than the MTJSR on different subsets.  However, the reported testing time of MTJSR in \cite{Celeb1000} is very high (3,254 seconds) on the subset-1000. In contrast, DELM only takes 350 seconds for training and 1.7 seconds for testing. Thus, compared to previous image set classification algorithms the proposed DELM based framework is more scalable to large scale datasets.

%\subsection{Robustness Experiments}
\begin{figure}
\begin{center}
   \includegraphics[width = 8.0cm]{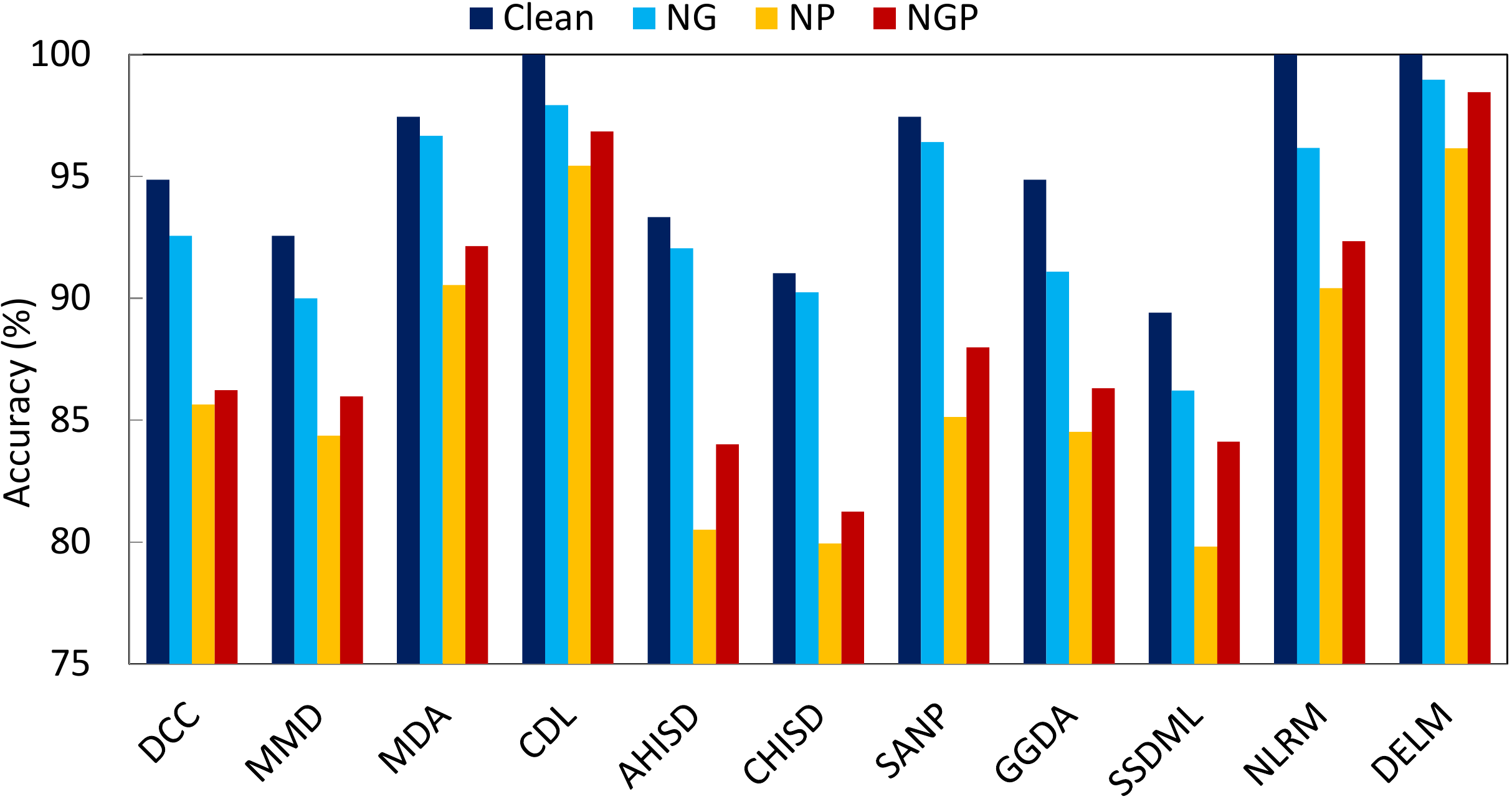}
\end{center}
\vspace{-4mm}
\caption{Average accuracy of different image set classification algorithms when the image-sets are corrupted by noise.}
\label{fig:accVnoise}
\vspace{-6mm}
\end{figure}
\noindent {\bf Robustness:} We tested the robustness of our algorithm to noise in a setup similar to \cite{AHISD, CDL}. From the Honda dataset, we generate clean data containing 100 randomly selected images per set in both the gallery and the probes to ensure that the same ratio of noise can be added to all sets. Image sets are then corrupted by adding one randomly selected image from each of the other classes. The original clean data and the three noisy cases are referred to as $N_c$ (clean), $N_G$ (only Gallery has noise), $N_P$ (only probe has noise) and $N_{G+P}$ (both gallery and probe have noise). Figure \ref{fig:accVnoise} shows that the proposed algorithm shows more robustness compared to other methods. As expected, sample based algorithms (AHISD, CHISD, SANP) are more sensitive to noise compared to the structure based ones, since modelling the set as a whole can resist the influence of noisy samples.
\begin{figure}
\begin{center}
   \includegraphics[width = 8.0cm]{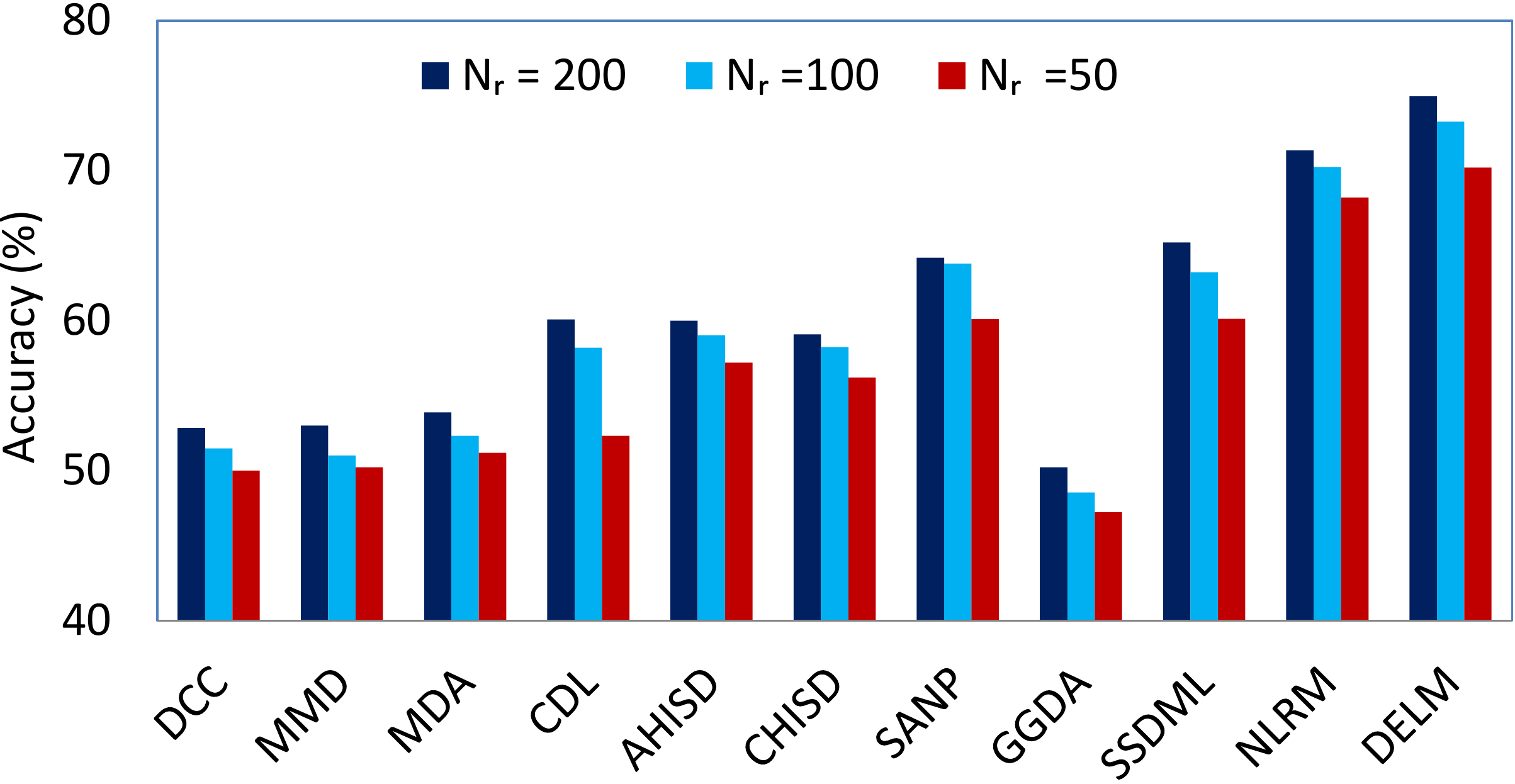}
\end{center}
\vspace{-4mm}
\caption{Robustness to the number of images per set. $N_r$ samples are randomly selected.}
\label{fig:accVsize}
\vspace{-2mm}
\end{figure}
We also perform evaluation with respect to varying numbers of samples per set. From the Youtube celebrities dataset, we randomly selected $N_r$ samples from each image set (both training and testing) and used them for recognition. In case there are less than $N_r$ samples in a set, we use all the samples. Figure \ref{fig:accVsize} shows the average accuracy of different methods for three values of $N_r$. The proposed algorithm is more robust and consistently outperforms other for decreasing value of $N_r$.

\vspace{1mm}
\noindent {\bf Execution Time:} We compare execution times on the Youtube celebrities dataset. Table~\ref{tab:time} shows the average execution times over the 5-fold experiments using a Core i7 3.4GHz CPU with 8GB RAM running MATLAB. The proposed algorithm is significantly faster than the existing state of the art in both training and testing. For example, our method takes only 1.01 seconds in training compared to 6542 seconds for NLRM, while achieving better accuracy.

%\begin{figure}
%\begin{center}
%   \includegraphics[width = 8.0cm]{./figures/ExecutionTime}
%\end{center}
%\label{tab:time}
%\vspace{-10mm}
%\end{figure}

\begin{table}[t]
\footnotesize
\caption{Execution times (in seconds) and training memory requirements (in megabytes) on the Youtube celebrities data. Test time is for matching one probe set to 141 gallery sets.}
\begin{center}
\vspace{-0mm}
%\begin{tabular}{|l|r|r|r|}
\begin{tabular}{lrrr}
\hline
\hline
      Algorithm & Training (sec)  & Testing (sec) & Memory (MB) \\
\hline
\hline
    DCC\cite{DCC} & 167.49 & 8.08 & 20.8\\
%\hline
    MMD\cite{MMD} & 313.57 & 78.32 & 150.2\\
%\hline
    MDA\cite{MDA} & 580.70 & 201.48 &   $ > 4 \times 10^4 $\\
%\hline
    AHISD\cite{AHISD} & -  & 18.10 & 93.7 \\
%\hline
    CHISD\cite{AHISD}& - & 190.61 & 971.4 \\
%\hline
    SANP\cite{SANP} & - & 17.94 & 160.6 \\
%\hline
    CDL\cite{CDL} & 345.88 & 13.08 & 238.8 \\
%\hline
	GGDA\cite{GGDA} & 450.92 & 20.24 & 200.0 \\
%\hline
   SSDML\cite{SSDML} & 400.01 & 21.87 & 127.7 \\
%\hline
  NLRM\cite{NLRM} & 6542 & 0.54 & 523.7 \\
%\hline
 Proposed DELM & \textbf{1.01} & \textbf{0.06}  & \textbf {14.3} \\
\hline
\end{tabular}
\end{center}
\label{tab:time}
\vspace{-9mm}
\end{table}

\vspace{1mm}
\noindent {\bf Memory Requirement:}
We also compare the training memory size requirement of the proposed algorithm with other algorithms on the Youtube Celebrities dataset.  DELM has lower training memory requirements (14.3MB) to achieve better classification results compared to other image set classification algorithms(Table~\ref{tab:time}).

%\begin{table}[t]
%\footnotesize
%\caption{Execution times, in seconds, on the Youtube celebrities data. Test time is for matching one probe set to 141 gallery sets.}
%\begin{center}
%\vspace{-2mm}
%\begin{tabular}{|l|r|r|}
%\hline
%      Algorithm & Training   & Testing \\
%\hline
%    DCC\cite{DCC} & 167.49 & 8.08\\
%%\hline
%    MMD\cite{MMD} & 313.57  & 78.32\\
%%\hline
%    MDA\cite{MDA} & 580.70 & 201.48\\
%%\hline
%    AHISD\cite{AHISD} & -  & 18.10\\
%%\hline
%    CHISD\cite{AHISD}& - & 190.61 \\
%%\hline
%    SANP\cite{SANP} & - & 17.94\\
%%\hline
%    CDL\cite{CDL} & 345.88 & 13.08 \\
%%\hline
%	GGDA\cite{GGDA} & 450.92 & 20.24 \\
%%\hline
%   SSDML\cite{SSDML} & 400.01 & 21.87 \\
%%\hline
%  NLRM\cite{NLRM} & 6542 & 0.54 \\
%%\hline
% Proposed DELM & \textbf{1.01} & \textbf{0.06}  \\
%\hline
%\end{tabular}
%\end{center}
%\label{tab:time}
%\vspace{-7mm}
%\end{table}

%%%%%%%%%%%%%%%%%%%%%%%%%%%%%%%%%%%%%%%%%%%%%%%%%%%%%%%%%%%%%%%%
%% Conclusions
%%%%%%%%%%%%%%%%%%%%%%%%%%%%%%%%%%%%%%%%%%%%%%%%%%%%%%%%%%%%%%%%
\vspace{-2mm}
\section{Conclusion}
\vspace{-2mm}
We presented an algorithm for learning of the non-linear structures of image sets for efficient and accurate classification. Our algorithm does not make any assumptions about the underlying image-set data and is scalable to large datasets. Non-linear structure is learned with the Deep Extreme Learning Machines (DELM) that enjoy the very fast training times of ELMs while providing deeper representations. Moreover, DELM models can be accurately learned from smaller image sets containing only a few samples. Experiments on five benchmark datasets show that our algorithm consistently outperforms 10 existing state of the art methods in terms of accuracy and execution time.

\section{Acknowledgement}
This work was supported by the Australian Research Council (ARC) Grants DP1096801 and DP110102399 and UWA Research Collaboration Award 2014.
\balance

%\F{Refine bibliography to have consistent names} \U{Please check the revised bibliography}
%%%%%%%%%%%%%%%%%%%%%%%%%%%%%%%%%%%%%%%%%%%%%%%%%%%%%%%%%%%%%%%%
%% References
%%%%%%%%%%%%%%%%%%%%%%%%%%%%%%%%%%%%%%%%%%%%%%%%%%%%%%%%%%%%%%%%
{\small
\bibliographystyle{ieee}
\bibliography{CVPRbib}
}

\end{document}